%% file: egpaper_for_review.tex
\documentclass[10pt,twocolumn,letterpaper]{article}

\usepackage{iccv}
\usepackage{times}
\usepackage{epsfig}
\usepackage{graphicx}
\usepackage{amsmath}
\usepackage{amssymb}
\usepackage[warn]{textcomp}

\usepackage{algorithm,algpseudocode}

\iccvfinalcopy 

\def\iccvPaperID{1917} 

\ificcvfinal\fi
\begin{document}

\title{Densely tracking sequences of 3D face scans}

\author{Huaxiong DING\\
Ecole Centrale de LYON\\
{\tt\small huaxiong.ding@ec-lyon.fr}
\and
Liming Chen\\
Ecole Centrale de LYON\\
{\tt\small liming.chen@ec-lyon.fr}
}

\maketitle

\input{Sections/Abstract}

\input{Sections/Introduction}
\input{Sections/RelatedWorks}
\input{Sections/ProposedMethod}

\input{Sections/Experiments}
\input{Sections/Conclusion}

This work was supported in part by the French Research Agency, l'Agence Nationale de Recherche (ANR), through the Jemime project (\textnumero contract ANR-13-CORD-0004-02), the Biofence project (\textnumero contract ANR-13-INSE-0004-02) and the PUF 4D Vision project funded by the Partner University Foundation.

{\small
\bibliographystyle{ieee}
\bibliography{egbib}
}

\end{document}

%% file: Sections/Abstract.tex
\begin{abstract}

3D face dense tracking aims to find dense inter-frame correspondences
in a sequence of 3D face scans and constitutes a powerful tool for
many face analysis tasks, \textit{e.g.}, 3D dynamic facial expression
analysis. The majority of the existing methods just fit a 3D face
surface or model to a 3D target surface without considering temporal
information between frames. In this paper, we propose a novel method
for densely tracking sequences of 3D face scans, which extends the
non-rigid ICP algorithm by adding a novel specific criterion for
temporal information.  A novel fitting framework is presented for
automatically tracking a full sequence of 3D face scans. The results
of experiments carried out on the BU4D-FE database are promising,
showing that the proposed algorithm outperforms state-of-the-art
algorithms for 3D face dense tracking.

\end{abstract}

%% file: Sections/Introduction.tex
\section{Introduction}

In facial behavior analysis, especially facial expression recognition,
the use of 3D face scans is regarded as a promising
solution~\cite{sandbach2012static}. Because compared with 2D images
based system, it is relatively insensitive to facial pose changes,
illumination conditions, and other changes in facial appearance like
facial cosmetics. Moreover, a 3D facial scan contains more information
displayed by face than a single-view 2D image, as it can additionally
record out-of-plane changes of the face surface. Thus, 3D face based
systems for facial expression analysis have attracted many researchers
in the past decades.

\begin{figure}[t]
\begin{center}
  \includegraphics[width=0.8\linewidth]{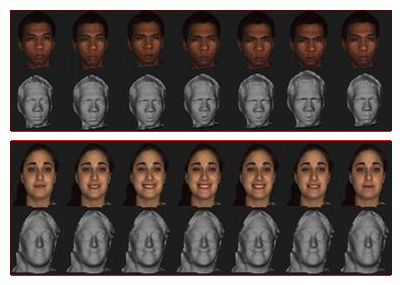}
\end{center}
   \caption{Examples from 3D dynamic facial expression databases BU4D-FE.}
   \label{fig:bu4d}
\end{figure}

Facial expression is naturally caused by the motions of facial muscles
beneath the skin of the face.  Consequently, the use of dynamic face
sequences is thought to be more reasonable and promising than only
using a static image, since it can encode both spatial and temporal
deformation information at the same time.  However, to capture a
sequence of 3D face scans is indeed much difficult than capturing a 2D
video. But fortunately, recent advances in 3D face data acquisition
(\ie structured light scanning~\cite{beumier19993d}, photometric
stereo~\cite{woodham1980photometric} and multi-view
stereo~\cite{beeler2010high}) have made it a feasible task. As a
result, a number of dynamic 3D face databases have been proposed for
this issue, like BU4D-FE~\cite{yin2008high} and
BP4D-spontaneous~\cite{zhang2014bp4d}, as shown in
Fig.~\ref{fig:bu4d}. More recently, with the popular use of low-cost
range sensors, such as Kinect, facial expression analysis using low
resolution dynamic 3D data has also attracted the attention.

To extract features of motions in the same location or region on the
different frames of sequences, accurate alignment and tracking methods
become very important for facial expression systems. Normally, 3D face
fitting, which aims at finding an optimal warp between a template
surface (\ie mesh or cloud points) and a target surface, is a key step
to solve this issue. According to the number of correspondent points,
the majority of existing methods in 3D face fitting can be divided
into two categories: sparse vs. dense. Unlike the former which only
locates a set of fiducial points (\eg eye corner and nose tip), the
latter needs to find a mapping from each point in the template onto
the target. Obviously, dense fitting is more challenging but more
attractive, as it is able to grab variations of 3D face scans more
accurately. Meanwhile, another advantage of dense fitting is that it
is easy to train and construct a statistical model based on the fitted
results and the established correspondences.

Dense face fitting has many applications in 3D face analysis, as an
accurate fitting algorithm is able to benefit facial
recognition~\cite{passalis2007intraclass, kakadiaris2007three} and
facial expression recognition~\cite{amberg2008expression,
  sandbach2012static, sandbach2012recognition}.  Normally, 3D dense
fitting can help align face scans and model facial expressions in the
majority of existing works~\cite{sandbach2012static}. More recently,
since 3D dense tracking can provide dense trajectories through
sequences of facial expression variations, improving facial expression
recognition using dense trajectory is another research
trend~\cite{afshar2016facial}.

Reviewing the development in 3D face dense fitting, various efforts
have been made for this issue, such as free-form deformations
(FFDs)~\cite{sandbach2011dynamic}, statistical
models~\cite{amberg2008expression, munoz2009direct}, non-rigid ICP
based algorithms~\cite{amberg2007optimal, cheng2016statistical},
harmonic maps~\cite{wang2008high}, and conformal
mappings~\cite{gu2008computational}.  However, the majority of
existing methods is initially proposed for static face scans fitting,
which means it just maps a template surface or model to a target
surface independently. Most of them disregard temporal information
between frames when they are used to fit a sequence rather than a
single face scan.

In this paper, we address the problem of 3D dense tracking and propose
a novel method for tracking sequences of 3D face scans. The
contributions are as follows: Firstly, we improve the non-rigid
Iterative Closest Point (ICP) algorithm by introducing a specific
criterion with respect to temporal information between frames. Besides
it, a novel dynamic framework is proposed to automatically fit full
sequences of 3D face scans iteratively. Moreover, we propose a joint
2D and 3D framework to detect facial landmarks on 3D face
scans. Furthermore, we present several techniques for non-rigid ICP to
improve the convergence and achieve a good initialization, including
employing a 3D identity/expression separated morphable model to
generate a template surface as a rough shape estimation and
introducing a variant of ICP with a scale factor to automatically
align two surfaces and rescale them into the same range.

The rest of the paper is organized as follows. Section 2 reviews 3D
dense fitting and tracking methods. Section 3 presents in detail the
proposed method including preprocessing, landmark detection and
fitting procedure. Section 4 shows the fitting results and discusses
some quantitative analysis in the experiments. Finally, Section 5
concludes this paper.

%% file: Sections/RelatedWorks.tex
\section {Related works}

Many efforts have been made to tackle the problem of 3D dense fitting,
for example, morphable models~\cite{munoz2009direct,
  amberg2008expression}, non-rigid ICP based
algorithms~\cite{allen2003space, amberg2007optimal,
  cheng2016statistical}, harmonic maps~\cite{wang2008high} and
conformal mappings~\cite{gu2008computational}. In this section, we
introduce different kinds of 3D face dense fitting algorithm and
discuss their advantages and disadvantages respectively.

Harmonic maps were firstly introduced by Wang
\etal~\cite{wang2008high} to find dense correspondences between 3D
face surfaces. The main idea of the proposed method is using harmonic
maps to project each surface to the canonical unit disk on the plane
with minimal stretching energy and bounded angle distortion. A sparse
set of easily detectable motion representative feature corners on the
disks is employed to find dense correspondence on 2D disk.  Conformal
mapping proposed by Gu \etal~\cite{gu2008computational} also projects
3D surface into a 2D plane, but it preserves angles between edges of
the mesh in the projection. The dense correspondence on 2D disk is
found by minimizing the matching function, which combines the optimal
Möbius transformation and the global matching function (OMGMF).  Both
harmonic map and conformal mapping are one-to-one, so the dense
correspondences on the 3D mesh can be established once the set of
correspondences on the 2D disks has been calculated.  Overall, the
advantage of conformal mapping is that it can provide very accurate
dense correspondence, but it requires meshes with high quality, needs
a prior knowledge of boundaries of surfaces, and it is very
computationally expensive.

3D morphable models (3DMMs) is a very useful technique that has been
exploited in 3D fitting and alignment. The fitting function normally
allows rigid deformations, such as rotation and translation, from the
mean shape, and the non-rigid transformation is defined as a linear
combination of basis vectors. Munoz \etal~\cite{munoz2009direct}
proposed a 3DMMs based method for tracking 2D image sequences, in
which the efficiency is achieved by factoring the Jacobian and Hessian
matrices appearing in the GaussNewton optimization into the
multiplication of a constant matrix depending on target structure and
texture, and a varying matrix that depends on the target motion and
deformation parameters. Amberg \etal~\cite{amberg2008expression}
proposed an iterative method for 3D face dense fitting, which used an
identity/expression separated 3D Morphable Model to fit the
correspondences found by ICP algorithm. Generally, 3DMMs based fitting
method can be very fast, and it is also robust to noise in the raw
input.  However, since the non-linearity of 3DMMs is ensured by PCA
decomposition basis, the fitting results are restricted by variations
of shape data used to build it. For example, a 3DMM established with
all faces with neutral expression is hard to fit a target face with
high-intensity expression. And besides, the complexity of local
geometry detail of the fitting results also indeed relies on the
number of PCA basis.

\begin{figure*}[ht]
\begin{center}
  \includegraphics[width=0.8\linewidth]{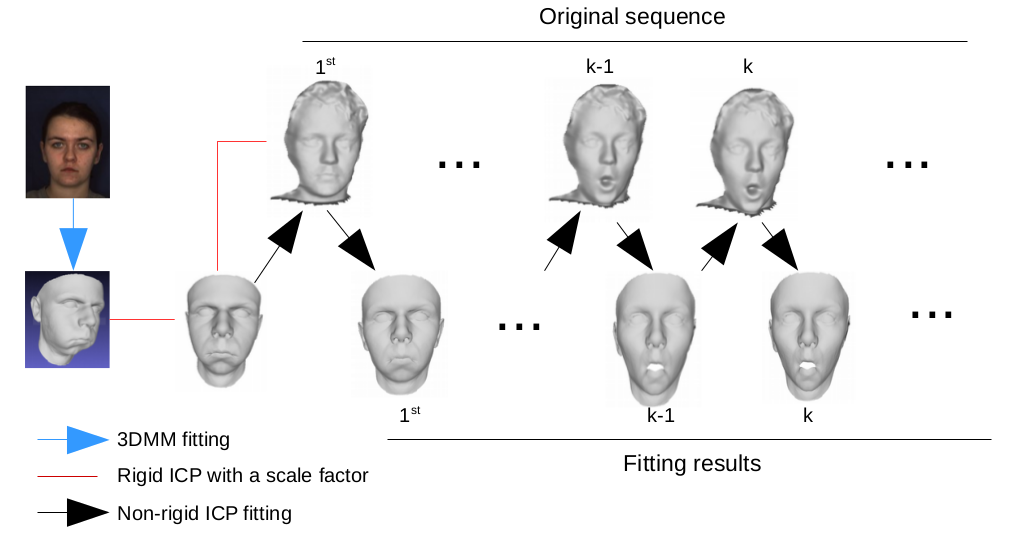}
\end{center}
   \caption{Overview of the proposed framework for automatically
     densely fitting sequences of 3D face scans}
\label{fig:overview}
\end{figure*}

Iterative close point (ICP) is another algorithm widely employed for
3D face alignment, which fits a template surface to a target
surface. Firstly, all points or a subset of points in the template are
designed to find a correspondent point in the target with the closest
distance. Then it wraps the template surface into the target surface
using the deformation computed from the correspondences. The
deformation of traditional ICP is rigid. Some
algorithms~\cite{allen2003space, amberg2007optimal,
  cheng2016statistical} that extend ICP methods to non-rigid
deformations have been proposed for 3D dense fitting. Amberg
\etal~\cite{amberg2007optimal} defined an affine deformation matrix
for each vertex in the template, and used a stiffness term to
regularize the non-rigid deformation.  More recently, Cheng
\etal~\cite{cheng2016statistical} proposed a novel method additionally
introducing the statistical prior as an extra constraint on the
fitting procedure of non-rigid ICP, in which a face mesh is first
divided into several regions by annotated model of the face
(AFM)~\cite{kakadiaris2005multimodal}, then on each subdivided region,
a combination of a linear statistical model and the non-rigid ICP is
employed to fit this local region. However, non-rigid ICP is able to
handle pose variations, occlusions, and even missing data. But it is a
little sensitive to initialization and vulnerable to noisy data as it
will fit to all points.

%% file: Sections/ProposedMethod.tex
\section{Proposed Method}

In this section, we present in detail the proposed method that is able
to automatically densely track sequences of 3D face scans.

Fig.~\ref{fig:overview} shows an overview of the framework employed in
the proposed method. First of all, a set of spatial facial landmarks
is located by a joint 2D and 3D face framework, which is used as a
prior knowledge for both 3DMMs and ICP fitting. Meanwhile, considering
that a good initialization can indeed boost the convergence of
ICP-like algorithms, we use a 3D morphable model to fit the $1_{st}$
frame of the sequences and adapt the fitted 3D face as a rough shape
estimation of $1_{st}$ frame, which will play a role of the template
surface in the ICP fitting process. Normally, the coordinates of the
fitted 3D face and the target surface are not in the range. Therefore,
a variant of rigid ICP with a scale factor is employed to coarsely
align and re-scale two surfaces to be fitted. After we have a coarse
alignment of that two surface, the proposed novel variant of non-rigid
ICP algorithm is used to refine local deformations. Iteratively, given
the fitting results of the previous frame, we use it as the template
surface to fit the face scan in the current frame. Finally, we repeat
the rule to fit a whole sequence until it ends.

\subsection{Joint 2D and 3D facial landmark detection}

Due to the importance of correspondence for the ICP-like algorithm,
our method reasonably relies on facial landmarks extracted from both
two face surfaces to be fitted, which are treated as a solid
correspondence to guide the convergence and restrict the large global
deformations.

\begin{figure}[t]
\begin{center}
  \includegraphics[width=0.8\linewidth]{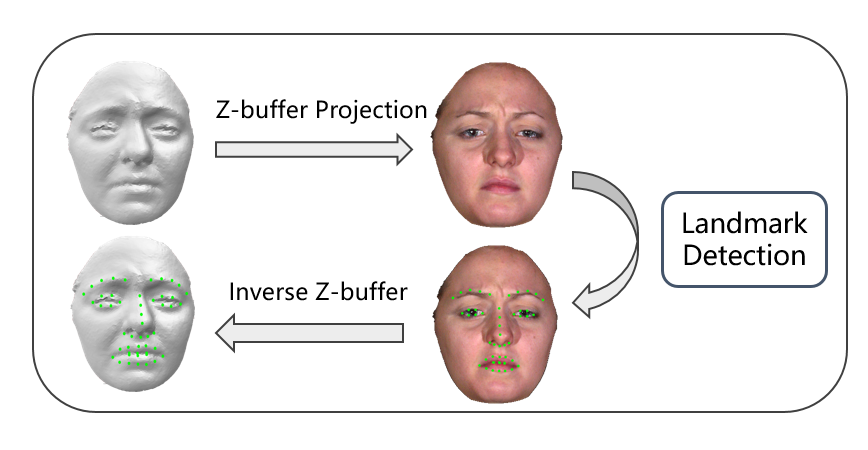}
\end{center}
   \caption{The joint 2D and 3D face landmark localization framework.}
   \label{fig:mark}
\end{figure}

Benefit from the recent breakthrough of the 2D face landmark detection
algorithm, detecting face landmarks on texture images has become more
reliable.  Here, we present a framework as described in
\cite{li2015efficient} using 2D landmark localization algorithm to
help locate face landmarks on the 3D mesh. More specifically, as shown
in Fig.~\ref{fig:mark}, a 3D face scan with texture map is firstly
projected into a 2D plane using z-buffer algorithm. Then a landmark
fitting algorithm based on an ensemble of regression
trees~\cite{kazemi2014one} is used to locate landmarks on the
projected 2D image. These 2D landmarks are further transferred to 3D
texture face space by the inverse of the above projection. Since the
one-to-one correspondence between 3D texture and 2D texture is
approximately preserved during the projection mapping, 3D landmarks
can be directly determined by the one-to-one correspondence between 3D
texture and 3D geometry of the 3D face model.

\subsection{3D Morphable model fitting}

\begin{figure}[t]
\begin{center}
  \includegraphics[width=0.8\linewidth]{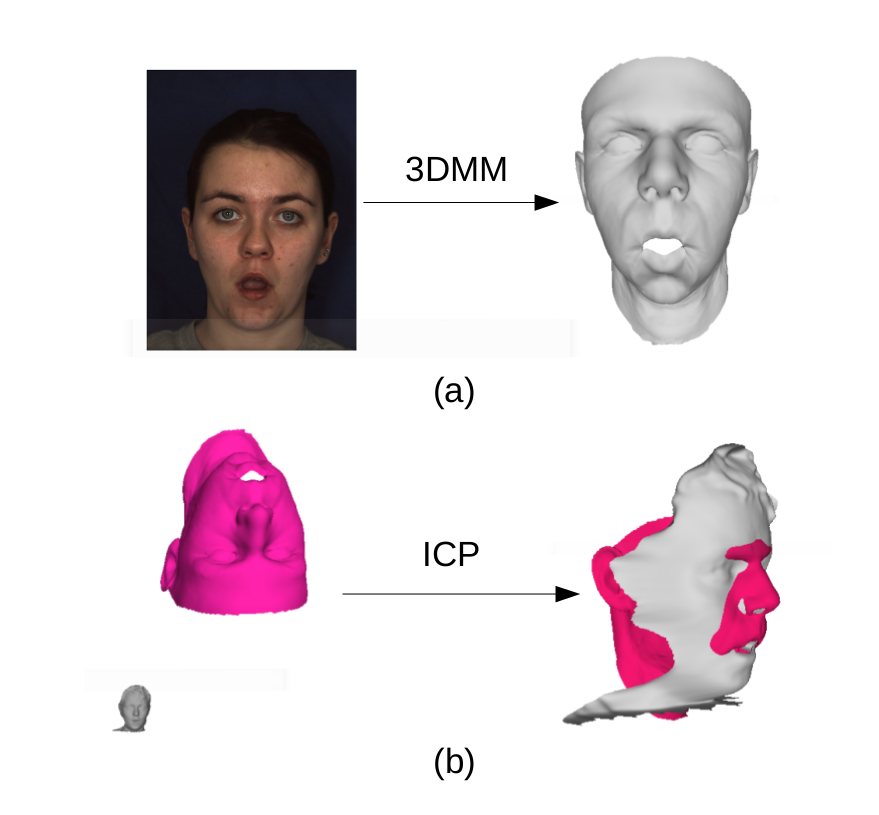}
\end{center}
   \caption{(a). A 3D face is fitted by a 3D morphable model from 2D
     texture images. (b). An ICP registration with a scale factor is
     used to align and re-scale two face surfaces.}
   \label{fig:3dMM}
\end{figure}

3D morphable model originally proposed by~\cite{blanz1999morphable} is
an effective method to describe 3D face space with the PCA. As
introduced previously, the non-linearity of 3DMMs is restricted by the
combination of PCA basis vectors.  Thus, to better fit the face with
various expressions, we introduce an identity/expression separated
morphable model~\cite{zhu2015high} to handle shape deformations caused
by variations of facial expression, \eg mouth open, as shown in
Fig.~\ref{fig:3dMM}.

\begin{equation}
  S(\alpha_{id}, \alpha_{exp}) = \bar{S} + A_{id}\alpha_{id} + A_{exp}\alpha_{exp}
\end{equation}
where $S$ is the 3D face, $\bar{S}$ is the mean shape, $A_{id}$
represents the identity component trained on the offset between
neutral face scans and the mean shape and $\alpha_{id}$ is the
identity coefficient, $M_{exp}$ represents the expression component
trained on the offset between expression scans and neutral scans and
$\alpha_{exp}$ is the expression coefficient.

Then the 3D face scan is projected into 2D plane:

\begin{equation}
  s_{2d} = fPR(S+t_{3d})
\end{equation}
where $s_{2d}$ is the projected image on image plane, $f$ is the scale
factor, $P$ is defined as the orthographic projection matrix, $R$ is
the rotation matrix on 3D space and $t_{3d}$ is the translation
matrix.

The fitting process is to minimize the difference between projected 2D
image and the image to be fitted.
\begin{equation}
  \arg\min_{f,R,t_{3d}, \alpha_{3d}, \alpha_{exp}}{\|s_{2d_{truth}} - s_{2d}\|}
\end{equation}

\subsection{ICP registration with a scale factor}

Given the 3D face fitted by 3DMMs, referred as the template surface,
we need further align it to the target surface that is a frame of the
3D dynamic face sequence. Actually, the coordinates of vertices in the
template are not in the same range as that of vertices in the target,
as shown in Fig.~\ref{fig:3dMM}. To well solve this issue, we use a
variant of rigid ICP equipped with a scale
factor~\cite{zinsser2005point}. Meanwhile, since facial landmarks have
been firstly located on both two surfaces, we employ all pairs of
landmarks as the correspondences used to calculate deformations in the
process of ICP.

The fitting function minimizes the difference when we warp points
$a_i$ in the template to points $b_j$ in the target.
 
\begin{equation}
  (R^{*},t^{*},s^{*}) = \arg\min_{R,t,s}{\sum_{(i,j)\in C}{{\|b_j -
        sRa_i - t\|}^2}}
\end{equation}
where $R$ is the rotation matrix, $t$ is the translation vector, $s$ is the scale factor.

To solve this function, the center of mass $\bar{a}$ of the selected
data points and the center of mass $\bar{b}$ of the corresponding
model points are firstly computed according to

\begin{equation}
  \bar{a} = \frac{1}{\lvert C \rvert}\sum_{(i,j)\in C}{a_i}, \qquad
  \bar{b} = \frac{1}{\lvert C \rvert}\sum_{(i,j)\in C}{b_i}
\end{equation}

This problem is solved by computing the SVD of the matrix
\begin{equation}
  K = \sum_{(i,j)\in C}{(b_j - \bar{b})(a_i - \bar{a})^T} = UDV^T
\end{equation}

where $R^* = UV^T$ that is regarded as an approximation of the
rotation matrix. Then the scale factor $s^*$ can be estimated by

\begin{equation}
  s^* = \frac{\sum_{(i,j)\in C}{\tilde{b_j}^T\tilde{a_i}}}{\sum_{(i,j)\in C}{\tilde{a_i}^T\tilde{a_i}}}
\end{equation}

where

\begin{equation}
  \tilde{b_j} = (b_j - \bar{b}),  \qquad
  \tilde{a_i} = R^*(a_i - \bar{a})
\end{equation}

Finally, the translation vector $t^*$ can be computed by $t^* =
\bar{b} - s^*R^*\bar{a}$.

\subsection{Non-rigid ICP with motion information}

In this section, we present in detail the proposed non-rigid ICP
algorithm that mainly extends the work~\cite{amberg2007optimal} with a
novel temporal information regularization.

We denote the template surface $\mathcal{S} = (\mathcal{V},
\mathcal{E})$, with a set of $n$ vertices $\mathcal{V}$ and a set of
$m$ edges $\mathcal{E}$. The target surface is denoted as
$\mathcal{T}$. The goal of fitting is to find a optimal mapping
$\mathcal{V}(X)$ wraping the template into the target while preserving
the topology structure.

More specifically, for each vertex in the template, we define a
$3\times4$ affine deformation matrix $X_i$ to allow the non-rigid
transformation. Therefore, we need to find an optimal $4n\times3$ matrix
$X$ to minimize the cost function with multiple regularizations in the
fitting process.

\begin{equation}
  X := {[X_1 \cdots X_n]}^T
\end{equation}

\subsubsection{Cost function}

The cost function used in this paper is similar to the one defined in
~\cite{amberg2007optimal}.  The difference is, that we add a new
specific item with respect to the motion history in previous frames to
meet the acquirement of tracking a sequence. Besides it, since we have
already automatically located facial landmarks on both two surfaces,
we use the item measuring the difference from the set of landmarks to
guide the convergence and restrict illegal deformations,
unlike~\cite{amberg2007optimal} uses manually annotated landmarks.

The proposed cost function consists of four items: a vertex distance
item, a stiffness term, a landmark distance item, and a motion history
item.

\textbf{Vertex distance item}: The item measures the Euclidean
distance between source vertex $v_i$ and its closest vertex $u_i$ in
the target surface.

\begin{equation}
  E_d(X) := \sum_{v_i \in \mathcal{V}}{w_i{\|X_iv_i - u_i\|}^2}
\end{equation}

where $w_i$ is a coefficient to weight the reliability of
correspondences. The robustness is achieved by setting $w_i$ to zero,
if it can not find the corresponding vertex in the case of missing
data.

\textbf{Stiffness item}: The item penalizes the differences between
the transformation matrices assigned to neighboring vertices that
belong to the same edge.  In fact, this item is employed to smooth the
deformation and avoid abrupt shape changes.

\begin{equation}
  E_s(X) := \sum_{\{i,j\} \in \mathcal{E}}{\|(X_i - X_j)G\|}^2_F
\end{equation}

where $G := diag(1, 1, 1, \lambda)$, $\lambda$ is used to weight
influences of the rotational and skew part against that of the
translational part in the deformation.

\textbf{Landmark item}: Similar to the regularization on vertices, the
item measures the distance between all pairs of landmark
correspondences $\mathcal{L} = \{ (v_1, l_1), \dots, (v_k, l_k) \}$.
\begin{equation}
  E_l(X) := \sum_{(v_i,l) \in \mathcal{L}}{\|X_iv_i - l_i\|}^2
\end{equation}

\textbf{Motion history item}:As introduced previously, the one-to-one
correspondences are naturally established in the set of fitted face
scans. Therefore, for each vertex, we could obtain a motion trajectory
through a sequence.  Based on techniques of trajectory analysis, an
approximation of the position of a vertex in the current frame can be
estimated according to the positions of that point in the previous
frames.

Following this idea, we could generate a face surface using these
estimated coordinates of vertices, then use it as a regularization
factor to guide the transformation in the fitting process.

The definition of the item is as follows:
\begin{equation}
  {v_i^k}^* = f(v_i^1, \dots, v_i^{k-1})
\end{equation}
where ${v_i^k}^*$ is a estimated vertex, and $f$ is a trajectory
estimator. In the experiments, we only use the classic Kalman
estimator as it is easy to be implemented. Different kinds of more complex
trajectory estimators can be found
in~\cite{barfoot2014batch,gibson2016optical}.

Similarly, the item also take the distances of vertices as the
measurement.
\begin{equation}
  E_m(X) :=  \sum_{v_i \in \mathcal{V}}{{\|X_iv_i - v_i^*\|}^2}
\end{equation}

\textbf{Cost function}: The full cost function is a weighted sum of these terms:
\begin{equation}
  E(X) := E_d(X) + \alpha E_s(X) + \beta E_l(X) + \gamma E_m(X)
\end{equation}

Here, $\alpha$ is the stiffness weight that influences the flexibility
of the template, $\beta$ weights the importance of facial landmarks,
and $\gamma$ is related to the influence of motion information.

\subsubsection{Parameters optimization}

We follow a similar framework as~\cite{amberg2007optimal} to find the
optimal affine transformation by minimizing the cost function
iteratively. For each iteration, the preliminary correspondences are
firstly determined. Then, based on the preliminary correspondences, we
calculate the optimal transformation parameters for the current
iteration. The transformed template and the target will lead to
finding the new correspondences for next iteration.  This process is
repeated until it converges, as shown in Algo.~\ref{algo:icp}.

\begin{algorithm}
  \caption{Non-rigid ICP fitting algorithm}
  \label{algo:icp}
  \begin{algorithmic}[1]
    \Require
    \Statex Two 3D face meshes: template $\mathcal{V}$ and target $\mathcal{T}$
    \State Initialize affine parameters $X^0$
    \State Choose a set of stiffness coefficients $\{\alpha^1, \dots, \alpha^n\}$, $\alpha_i > \alpha_{i+1}$,
     landmark coefficients $\{\beta^1, \dots, \beta^n\}$, $\beta_i > \beta_{i+1}$
    and motion coefficients $\{\gamma^1, \dots, \gamma^n\}$, $\gamma_i > \gamma_{i+1}$.
    \For { each $\alpha_i$, $\beta_i$, $\gamma_i$}
    \While { $\| X_j - X_{j-1} \| > \varepsilon$}
    \State Find preliminary correspondences for $\mathcal{V}(X)$
    \State Calculate optimal local affine transform $X_j$ 
    \State based on correspondences, $\alpha_i$, $\beta_i$ and $\gamma_i$
    \EndWhile
    \EndFor

    \State \Return $X$
    
  \end{algorithmic}
\end{algorithm}

Actually, once the correspondences are fixed, the cost function can be
transferred to a sparse quadratic system which can be minimized
exactly. More specially, in order to differentiate the formulation, we
rewrite items of cost function into canonical form. For the vertex
distance item, we swap the positions of the unknowns and the fixed
vertices:
\begin{equation}
  E_d(X) := {\| W(DX - U) \|}_F^2
\end{equation}

Where $W := diag(w_1, \dots, w_n)$, D is a sparse matrix mapping the
$4n\times3$ matrix of unknowns $X$ onto displaced source vertices.

For the stiffness item, we define a node-arc incidence matrix $M$. If
edge $r$ connects the vertices $(i, j)$ and $i<j$, the nonzero entries
of $M$ in row $r$ are $M_{ri} = −1$ and $M_{rj} = 1$. Then the item
can be rewritten as:

\begin{equation}
  E_s(X) := {\| (M \otimes G)X \|}_F^2
\end{equation}

Similarly, the rest two item can also be rewritten in the form of

\begin{equation}
  E_l(X) := {\| D_LX - U_L \|}_F^2
\end{equation}

\begin{equation}
  E_m(X) := {\| DX - U_m \|}_F^2
\end{equation}

Now, the original cost function becomes a quadratic function:

\begin{equation}
  \begin{array}{lcl} 
    E(X) & = & {\left\Vert \left[ \begin{array}{c} WD \\ \alpha M \otimes G \\ \beta D_L \\ \gamma D \end{array} \right] X - \left[ \begin{array}{c} WU \\  0 \\  \beta U_L \\ \gamma U_m \end{array} \right] \right\Vert_F^2} \\
     
    & & \\
    
  & = & \left\Vert AX - B \right\Vert_F^2
  \end{array} 
\end{equation}

which is a typical linear least square problem. And $E(X)$ takes on its
minimum at $X = (A^TA)^{-1}A^TB$.  Thus, For each iteration, given
fixed correspondences and coefficents, we could determine the optimal
deformation quickly.

%% file: Sections/Experiments.tex
\section{Experimental Evaluation}

In this section, we present our experiments and discuss the
experimental results in detail.

\subsection{Database}
BU4D-FE~\cite{yin2008high} is a high-resolution 3D dynamic facial
expression database, which contains facial expression sequences
captured from 101 subjects. For each subject, there are six model
sequences showing six prototypic facial expressions (anger, disgust,
happiness, fear, sadness, and surprise), respectively. Each expression
sequence contains about 100 frames.  Each 3D model is equipped with a
high-resolution texture map. Some examples have been shown in
Fig.~\ref{fig:bu4d}

\subsection{Experimental setup}

The experiments are carried out on a subset of BU4D-FE database. The
sequences of 20 subjects are selected to evaluate the proposed method.
Since ICP needs to find correspondences for each point on surfaces,
several preprocessings have been applied to remove noise points, such
as spikes and small isolated pieces, for achieving a robust
performance.

Now, we introduce some details for implementation and present
parameters that we used in the experiments. For landmark detection, we
employ the algorithm proposed in~\cite{kazemi2014one} and locate
totally 51 landmarks on the face. In rigid registration, all the
surfaces are aligned and re-scaled into a cube $[0,1]$ to avoid too
large value in the computation. In the stage of Non-rigid ICP, once
the correspondences have been established by the closest distance,
another criterion is applied to increase the reliability, which is
achieved by dropping correspondences if the angle between normal
directions of two correspondent points is large than $\frac{\pi}{4}$.
Finally, in the cost function of non-rigid ICP, $\alpha$ and $\beta$
vary from $100$ to $10$ with a step $10$, while $\gamma$ varies from
$5$ to $0.5$ with a step $0.5$;

To evaluate the proposed method and compare its fitting result with
the state of the art algorithm, we introduce two methods as the
baselines. The first method is proposed by ~\cite{amberg2007optimal},
referred as NICP, based on which we develop our proposed method. Since
3DMM have been widely used in modeling 3D facial expressions, and even
in dense tracking~\cite{munoz2009direct}, we adapt the 3DMM fitting as
our second method for comparison.

\subsection{Quantitative analysis}

Firstly, to measure the quality of registration, we propose a method
based on facial landmarks. More specially, if we have already located
facial landmarks on all 3D face scans, we could extract local geometry
properties, such coordinates and normal directions, in an $n$-ring
region around facial landmarks. Then the difference of extracted
geometry properties between the fitting result and the target surface
is regarded as the measurement of the registration quality.

Given a set of landmark $\{(v_1, l_1), \dots, (v_k, l_k)\}$, $f(x)$
represents the average value of local geometry properties of regions
around the landmark $x$.

\begin{equation}
  measurement : = \frac{1}{k}\sum_{i=1..k}D(f(v_i), f(l_i))
\end{equation}
where $D$ indicates the Euclidean distance for coordinates, and the
angle for normal directions.

\begin{table}
\begin{center}
\begin{tabular}{|l|c|c|}
\hline
Method & Mean Error & Mean Error \\
&  (coords $\times10^{-3}$) & (normals) \\
\hline\hline
3DMM & 11.3 & 0.41 \\
NICP & 9.9 & 0.32 \\
Our Method & 8.6 & 0.31 \\
\hline
\end{tabular}
\end{center}
\caption{Average error of local geometry properties computed from regions around landmarks.}
\label{table:error}
\end{table}

Table~\ref{table:error} shows the performance measures of the three
fitting methods. We can observe that both two non-rigid ICP based
methods outperform 3DMM based fitting algorithm. We think the
principle reason to the failure of 3DMM is that the non-linearity of
3DMM is limited by its PCA components and it can not capture
out-of-plane changes since it fits 2D images.

However, our method clearly outperforms NCIP in term of coordinate
error, which proves that the use of motion information can indeed
improve the accuracy and robustness of the fitting results. As a
result, it is able to provide more stable dense trajectories for 3D
dynamic expression analysis. Meanwhile, its error of normals is still
slightly lower than that of NCIP, showing the effectiveness of
proposed framework.

\subsection{Visualization of fitting results}

\begin{figure}[t]
\begin{center}
  \includegraphics[width=0.8\linewidth]{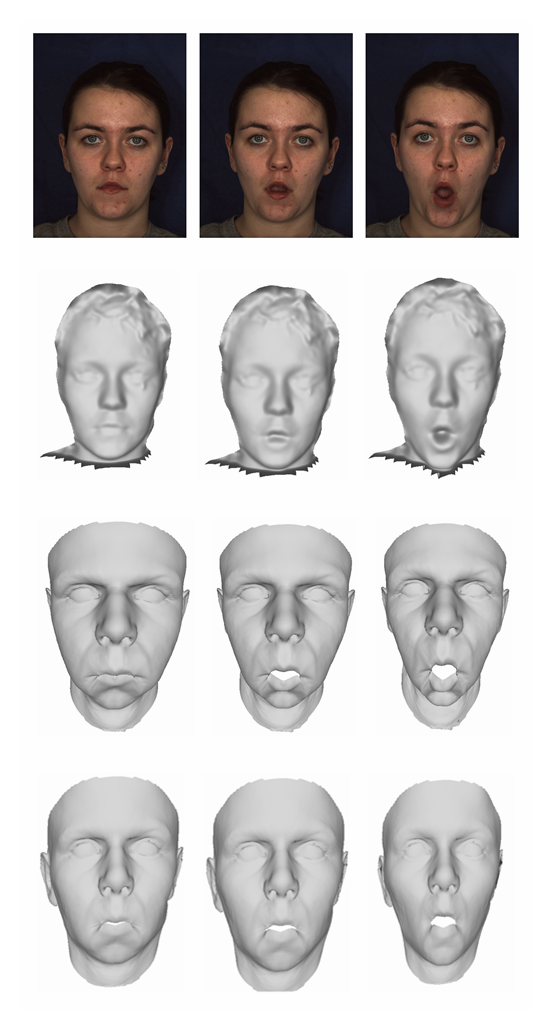}
\end{center}
   \caption{Fitting results for scans with different intensity of
     expressions from 3D dynamic facial expression databases
     BU4D-FE. The 1st row presents original texture images and the 2nd
     row shows original meshes. the 3rd row is the fitting result of
     3DMM and the 4th row presents the fitting results of our method.}
   \label{fig:result}
\end{figure}

We show some fitting results with the different intensity of
expressions in Fig.~\ref{fig:result}, from which we can observe that
the proposed model fits 3D face much better than 3DMM.  Actually, the
mouth open is still a challenge for most of the fitting algorithm. The
surface of 3D faces fitted by 3DMM is uneven, especially in the region
around the mouth, that may be caused by the limitation of
non-linearity as we discussed previously. The 3D faces fitted by the
proposed method display more details, such as eyelids in a smooth
surface.  Meanwhile, it is able to capture the deformation of mouth
caused by variations of expressions accurately, which also
demonstrates the power of the proposed method.

%% file: Sections/Conclusion.tex
\section{Conclusion and Future Work}

In this paper, we tackle the issue of 3D dense tracking and propose a
novel framework that is able to densely track the sequences of 3D face
scans. The techniques employed in the fitting process to achieve
robust results are introduced in detail, including landmark detection,
3DMM fitting, and two-stage ICP registration. To satisfy the terms of
dense tracking, we extend the non-rigid ICP by taking into a new
criterion into motion information. The experimental results prove that
the proposed method is promising to solve this issue.

The family of ICP algorithm relies heavily on the correspondence
establishing, which mostly determines the convergence. In this paper,
we find the correspondences between two surfaces only using geometry
information, such as coordinates and normal directions. Since 3D face
scans are usually along with the texture map, to make use of texture
information, the proposed methods~\cite{weinzaepfel2013deepflow}
designed for finding correspondences on 2D images can be integrated in
the future.